\documentclass[a4paper,twoside]{article}
\usepackage{microtype}
\usepackage{epsfig}
\usepackage{subcaption}
\usepackage{calc}
\usepackage{amssymb}
\usepackage{amstext}
\usepackage{amsmath}
\usepackage{amsthm}
\usepackage{booktabs}
\usepackage{multicol}
\usepackage{pslatex}
\usepackage{apalike}
\usepackage{graphicx}
\usepackage[hidelinks]{hyperref}
\usepackage{SCITEPRESS}     

\begin{document}

\title{Unsupervised Domain Adaptation from Synthetic to Real Images for Anchorless Object Detection }

\author{\authorname{Tobias Scheck\orcidAuthor{0000-0002-1829-0996}, Ana Perez Grassi\orcidAuthor{0000-0003-1171-903X}, Gangolf Hirtz\orcidAuthor{0000-0002-4393-5354}}
    \affiliation{Faculty of Electrical Engineering and Information Technology, Chemnitz University of Technology, Germany}
    \email{\{tobias.scheck,ana-cecilia.perez-grassi,g.hirtz\}@etit.tu-chemnitz.de}
}

\keywords{Unsupervised Domain Adaptation, Synthetic Images, CenterNet, Anchorless/Keypoint-based Detectors, Object Detection.}

\abstract{
    Synthetic images are one of the most promising solutions to avoid high costs associated with generating annotated datasets to train supervised convolutional neural networks (CNN).
    However, to allow networks to generalize knowledge from synthetic to real images, domain adaptation methods are necessary.
    This paper implements unsupervised domain adaptation (UDA) methods on an anchorless object detector.
    Given their good performance, anchorless detectors are increasingly attracting attention in the field of object detection.
    While their results are comparable to the well-established anchor-based methods, anchorless detectors are considerably faster.
    In our work, we use CenterNet, one of the most recent anchorless architectures, for a domain adaptation problem involving synthetic images.
    Taking advantage of the architecture of anchorless detectors, we propose to adjust two UDA methods, viz., entropy minimization and maximum squares loss, originally developed for segmentation, to object detection.
    Our results show that the proposed UDA methods can increase the mAP from $61\, \%$ to $69\, \%$ with respect to direct transfer on the considered anchorless detector.
    The code is available: \url{https://github.com/scheckmedia/centernet-uda}.
}
\onecolumn \maketitle \normalsize \setcounter{footnote}{0} \vfill

\section{\uppercase{Introduction}}
\label{sec:introduction}

Object detection, which involves both locating and classifying objects in an image, is one of the most challenging tasks in computer vision.
Its difficulty depends on the application, which can go from highly controlled environments with few well-known objects, such as in industrial tasks, to extremely complex and dynamic environments with a large number of varying objects, such as outdoor traffic scenes.

With the advent of CNNs (Convolutional Neural Networks), object detection has undergone a breakthrough. In this context, two important lines of work have recently appeared.
The first one aims to find the most efficient way to represent objects in order to train CNNs.
This has led to the development of keypoint-based (i.e., anchorless) detectors.
The second line of work addresses the increasing need of using training datasets that differ in nature from those of the real applications.
This is due to the great cost in generating and labeling large datasets from the real setting.
However, using such datasets that are different in nature to the intended ones poses the problem of domain adaption, i.e., translating from one dataset to the other.

Currently, object detection with CNNs is strongly dominated by anchor-based methods.
These methods involve very popular architectures like SSD \cite{fu_dssd_2017}, YOLO \cite{redmon_yolo9000:_2016}, R-FCN \cite{dai_r-fcn:_2016}, RetinaNet \cite{lin_focal_2017} and Faster R-CNN \cite{ren_faster_2017}.
To recognize an object, these networks generate thousands of RoIs (Regions of Interest) with different positions, sizes and shapes classifying them individually.
The number of RoIs is a design parameter and equal for all images, independent of the real number of objects that actually appear in them.
During the training, those RoIs that have a IoU (Interception over Union) value higher than 50\% with respect to the given ground truth are considered positives, while the rest are considered background.

On the one hand, the number of RoIs must be kept high in order to find all possible objects in the images.
On the other hand, this results in an imbalance, since normally the number of RoIs containing background will be much higher than those with objects.
Moreover, anchor-based methods require a careful design, where the adequate number, size, and shape of the RoIs depend on the application.

These disadvantages have motivated researchers to investigate on new architectures for object detectors. With Cornernet \cite{law_cornernet_2020}, a new generation of anchorless detectors has started to attract attention.
In this case, objects are represented by keypoints. The number of keypoints in one image is directly proportional to the number  of objects on it.
By eliminating the need of a fix and large amount of RoIs, design and imbalance issues are not only overcome, but also the resulting efficiency increases.
Our work is based on CenterNet \cite{zhou_objects_2019}, an anchorless architecture that represents objects using their center point and a dimension vector.

In practice, a big problem that all object detectors, anchor-based and anchorless, face, is the lack of sufficient labeled data to train the networks.
Not only capturing enough images, but specially labeling them, is very expensive.
For this reason synthetic images has been gaining in importance in this field.
Game engines allow not only generating synthetic images with realistic shapes, textures and movements, but they also label them automatically.
This way, CNNs can be trained on synthetic images to later work as detector on real ones.
In this case, synthetic images are the source domain and real images are the target domain.

However, the difference between real and synthetic images is still a problem for CNNs, which cannot generalize well from one domain to the other.
Different techniques has been developed to reduce the gap between target and source domain. In this work, we are interested in UDA (Unsupervised Domain Adaptation) methods, which use unlabeled target images to bring source and target domain closer.
We focus on two UDA techniques: entropy minimization \cite{vu_advent_2019} and maximum squares loss \cite{chen_domain_2019}.
These techniques were developed for segmentation applications, however, given the architecture of anchorless detectors, they can be easily adapted to object detection as shown in this paper.   \\

\noindent\textbf{Contributions. }This paper aims to translate UDA techniques from segmentation to object detection by taking advantage of the architecture of anchorless detectors.
We first test how an anchorless architecture, viz., CenterNet, generalizes from synthetic images to real ones and we compare its performance with anchor-based methods.
Then, we extend its architecture to consider UDA by entropy minimization and maximum squares loss.
We show that UDA methods can improve the performance of anchorless detectors trained on synthetic images.
To our best knowledge, this is the first work to test the anchorless detector CenterNet with synthetic images and extending it to consider UDA methods.
\section{\uppercase{State of the Art}}

With the advent of deep learning, object recognition have reached new state-of-the-art performances.
In the last years, architectures based on anchor boxes, like SSD \cite{fu_dssd_2017}, YOLO \cite{redmon_yolo9000:_2016}, R-FCN \cite{dai_r-fcn:_2016}, RetinaNet \cite{lin_focal_2017} and Faster R-CNN \cite{ren_faster_2017}, have dominated this field.

Despite of their success, the use of anchor boxes has considerable drawbacks.  Firstly, a large number of boxes is necessary  to ensure enough overlap with the ground truth. This latter results in much more negative than positive sample, i.e., in an unbalanced dataset impairing the training process. Secondly, the characteristics and number of the anchor boxes should be designed carefully and are normally customized for a given dataset or application.

To overcome these disadvantages, a new generation of anchorless detectors have been developed in the last year.
These detectors abandon the concept of anchor boxes and instead localize objects based on keypoints.
The pioneering anchorless detector is CornerNet \cite{law_cornernet_2020} -- note that this paper was available online since 2019. CornerNet describes an object as a pair of keypoints given by the top-left and bottom-right corners of a bounding box.
In \cite{duan_centernet_2019}, an extension of CornerNet is proposed by adding the center of the bounding box as keypoint.
In \cite{zhou_bottom-up_2019}, ExtremeNet is presented where keypoints are given by objects' extreme point.
Extreme points have the advantage over corners of being always part of the object, without being affecting by background information.

All the aforementioned detectors use more than one keypoint, therefore, it is necessary to group them in order to perform a detection on their basis.
For this reason, they are called point-grouping detectors \cite{duan_corner_2020}.
On the other hand, detectors called point-vector detectors \cite{duan_corner_2020} are based on only one keypoint and a vector containing geometrical characteristics of the object, like width, height, etc. In this category, we can enumerate CenterNet \cite{zhou_objects_2019}, FoveaBox \cite{kong_foveabox_2020} and FCO \cite{tian_fcos_2019}.

Our work uses CenterNet as defined in \cite{zhou_objects_2019}.
CenterNet models objects using their center point, which are detected from a CNN-generated heatmap.
From this keypoint, CenterNet is able to regress other object properties, such as size, 3D location, orientation and pose.
As CenterNet only uses one keypoint, it does not need any grouping stage, which makes it faster than the point-grouping detectors. Zhou et al. present four architectures for CenterNet using ResNet18, ResNet101 \cite{he_deep_2015}, DLA34 \cite{yu_deep_2019} and Hourglass-104 \cite{leibe_stacked_2016} as feature extractors. 

All the mentioned works have been tested on real image datasets, specifically, most of them use COCO dataset \cite{fleet_microsoft_2014}.
One of the most important disadvantages of supervised neural networks is their strong dependence on the quantity and variety of the training images as well as the quality of their labels.
This is one of the highest hurdle when implementing CNNs in real applications, since generating such a training dataset is associated with a huge effort and high costs.

To overcome this predicament, the use of synthetic images have been attracting attention in the last years.
Modern game platforms allow generating photorealistic images, introducing variations and modifications with less effort.
In addition, images are labeled automatically and without error, which reduces efforts even more.

In the area of autonomous driving, for example, synthetic dataset like GTA 5 \cite{leibe_playing_2016} and SYNTHIA \cite{ros_synthia_2016} are used to train neural networks for detection and segmentation tasks.
For indoor applications, the dataset SceneNet \cite{mccormac_scenenet_2017} and SceneNN \cite{hua_scenenn_2016} present segmentation masks, depth maps and point clouds of unoccupied rooms with different furniture.
SURREAL \cite{varol_learning_2017} combine real backgrounds and synthetic persons for human segmentation and depth estimation. THEODORE \cite{scheck_learning_2020} presents different indoor scenes from a top-view using an omnidirectional camera.

Despite the good quality of some synthetic images, their difference with real images, called reality gap, constitutes a problem for neural networks.
Tested on real images, neural networks trained with synthetic images have a worse result than those trained with real ones.
This problem has motivated an increasing investigation in the area of unsupervised domain adaptation (UDA) for synthetic images.
In this case, synthetic and real images constitute the source and target domain respectively.

UDA methods use unlabeled real images during the training to approximate the source domain to the target domain and therefore to minimize the reality gap.
In \cite{li_deep_2020} a survey of deep domain adaptation for object recognition is presented.
All works mentioned in \cite{li_deep_2020} involve, however, anchor-based architectures.
In this paper, in contrast to this, we are concerned with anchorless approaches.

For a segmentation task, Vu et al.~showed that models trained on only the source domain tend to produce low-entropy predictions on source-like (i.e., synthetic) images and high-entropy predictions on target-like (i.e., real) images \cite{vu_advent_2019}.
Based on this, the authors propose two methods using entropy minimization (EM) to adapt from the synthetic to the real domain.
One method minimizes the entropy indirectly by adversarial loss and the second one does it directly by entropy loss.
These methods are mainly applied on image segmentation, but also tested on object detection.
For object detection, the authors use a modified version of a SSD-300 \cite{leibe_ssd:_2016}.
Further, in \cite{chen_domain_2019}, Chen et al.\,observe that the gradient's magnitude of the entropy loss disproportionately favors those classes detected with high probability. This latter results in a training process dominated by those object classes that actually are easier to transfer from one domain to other.
To counteract this effect, Chen et al.\,propose to replace the entropy minimization by the maximum squares loss (MSL) \cite{chen_domain_2019}.
In this work, we adapt and compare these two UDA methods on anchorless detectors, more specifically on CenterNet.
In contrast to using SSD-300 as proposed in \cite{vu_advent_2019}, incorporating EM and also MSL in CenterNet can be done without altering its architecture.
\section{\uppercase{Background}}
%
In this section, we firstly introduce the detector CenterNet and both previously mentioned UDA methods, i.e., EM and MSL.
We then extend CenterNet's architecture to consider these UDA methods, in order to obtain an anchorless detector with domain adaptation between synthetic and real images.
\subsection{CenterNet}
CenterNet is an anchorless object detector describing objects as points, which was introduced in \cite{zhou_objects_2019}.
More specifically, CenterNet identifies each object using only the center point of its RoI.
Then, to regress the object size, CenterNet uses a vector with the RoIs' widths and heights. 

Let $\mathcal{C}=\{1,\dots, C\}$ be the set of all $C$ object classes to be detected.
The training dataset $\mathcal{T}$ is given by labeled images $\mathbf{x}_i$ of size $H \times W \times 3$, with $1\leq i \leq |\mathcal{T}|$.
Each object in an image $\mathbf{x}_i$ is annotated with a surrounding RoI and a class $c \in \mathcal{C}$, which together constitute the ground truth.

To train CenterNet, the ground truth should be converted from RoIs to heatmaps.
To this aim, for each image $\mathbf{x}_i \in\mathcal{T}$, a keypoint map $K(x,y,c)$ of size $H \times W \times C$, is generated by extracting the center points of each annotated RoIs.
$K(x,y,c)$ is equal one, only if the coordinates $(x,y)$ in $\mathbf{x}_i$ belong to an object's center of class $c$ and equal zero elsewhere.
Through convolution with a Gauss kernel, whose standard deviation is a function of the object's size, the keypoint map $K(x,y,c)$ is expanded to form a heatmap.
The size of this heatmap is then modified to agree with the size of the network's output by using a factor $R$.
The final heatmap of an image $\mathbf{x}_i$ is denoted by $Y(x,y,c) \in [1,0]$ and have a size of $\frac{W}{R} \times \frac{H}{R} \times C$.
The training dataset $\mathcal{T}$ and the corresponding set of heatmaps $Y(x,y,c)$ (new ground truth) are used together with a focal loss function $L_h$ \cite{lin_focal_2017} to train the network in order to predict new heatmaps $\hat{Y}(x,y,c)$.

The down-sampling of the heatmap using $R$ produces a discretization error at the location of the objects' center.
To correct this error, CenterNet also provides an output $\hat{O}\in\mathbb{R}^{\frac{W}{R} \times \frac{H}{R} \times 2}$ with the predicted offset.
Additionally, for each detected object's center, CenterNet also regresses the object's size, in order to reconstruct its RoI.
The predicted size is given by the output $\hat{S}\in \mathbb{R}^{\frac{W}{R} \times \frac{H}{R} \times 2}$.
Offset and size outputs are trained using the $L_1$-loss functions $L_{off}$ and $L_{size}$ respectively.
Finally, the linear combination of all loss functions, i.e., from heatmap, offset and size gives the complete detection loss function of CenterNet,where $\lambda_{h}$, $\lambda_{size}$ and $\lambda_{off}$ are scale factors \cite{zhou_objects_2019}:
\begin{eqnarray}
    \label{eq:detection_loss}
    L_{det}(\mathbf{x}_i)=\lambda_{h}L_h+\lambda_{size}L_{size}+\lambda_{off}L_{off}.
\end{eqnarray}

At inference time, the 100 highest peaks inside $8$ neighborhoods are extracted from the predicted heatmap $\hat{Y}(x,y,c)$.
The coordinates of each of these peaks may indicate the center of a detected object.
The probability of each detection, given by the corresponding value of $\hat{Y}(x,y,c)$, is used as a threshold to validate the detection.
To reconstruct the RoIs of the detected objects, the coordinates of the center points in $\hat{Y}(x,y,c)$ are corrected using $\hat{O}(x,y)$, while the width and height dimensions are extracted from $\hat{S}(x,y)$.

The architecture of CenterNet consists of one feature extractor followed by three heads, one for each of the described outputs: heatmap head, offset head and size head. 
In \cite{zhou_objects_2019}, four architectures are presented as feature extractor: ResNet18, ResNet101 \cite{he_deep_2015}, DLA34 \cite{yu_deep_2019} and Hourglass-104 \cite{leibe_stacked_2016}.
We choose ResNet101 and DLA34 for our experiments because they present the best trade-off between accuracy and runtime \cite{zhou_objects_2019}.


%
\begin{figure*}[!h]
    \centering
    \includegraphics[width=.8\textwidth]{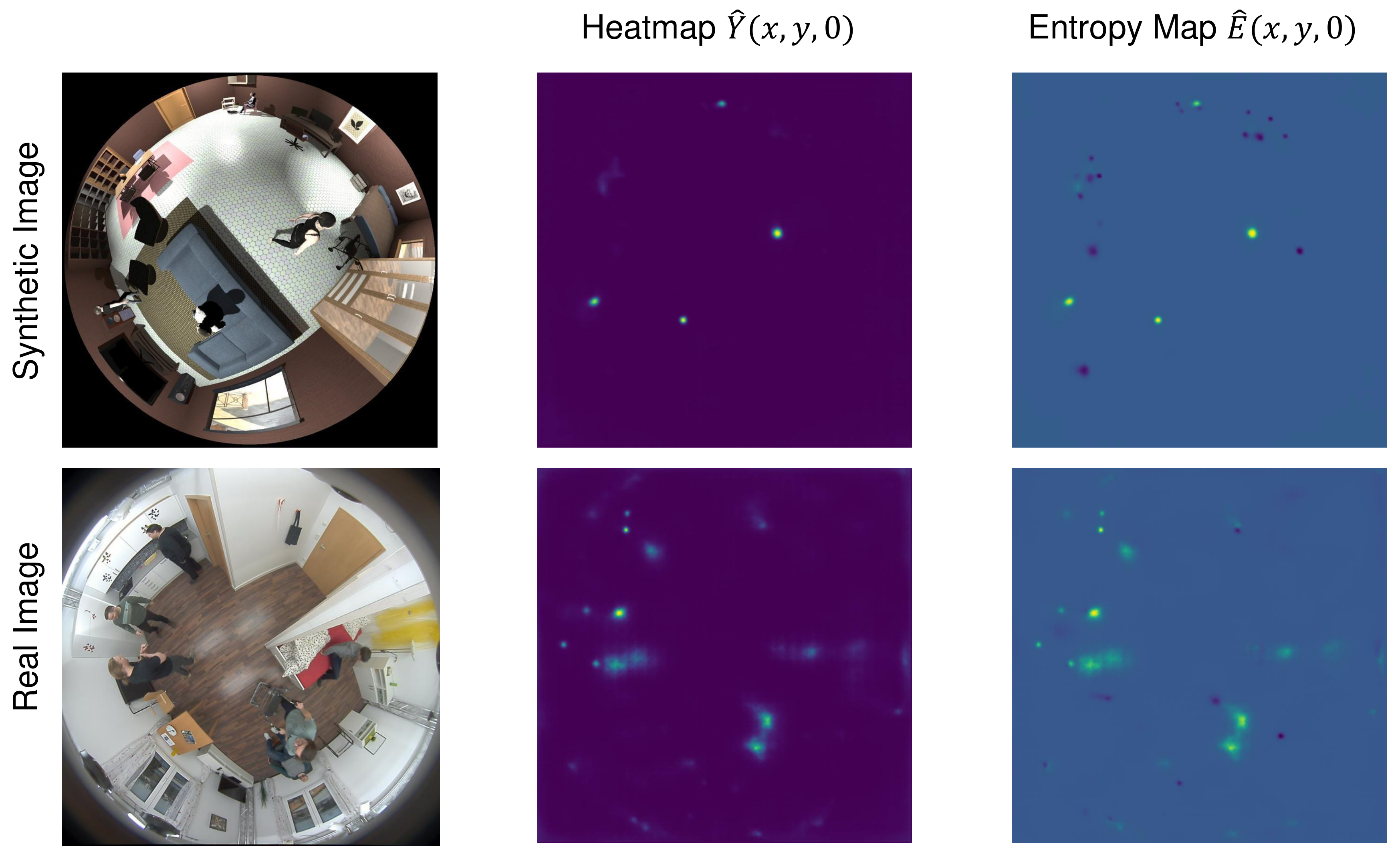}
    \caption{Heatmap and entropy map for $c=0$ (person) from source (i.e., synthetic) and target (i.e., real) image. Note: For ease of exposition,  the entropy map is shown only for one class.}
    \label{fig:HeatmapEntropy}
\end{figure*}

\subsection{UDA by Entropy Minimization}
\label{Sec:UDA}


In \cite{vu_advent_2019} it was shown that segmentation models trained only with synthetic images (source domain) tends to produce low-entropy predictions on other synthetic images, but high-entropy predictions on real images (target images).
Based on this observation, it is possible to reduce the gap between synthetic and real images by enforcing low-entropy predictions on real images.

In a segmentation architecture, the prediction output for one image $\mathbf{x}_i$ consists of a segmentation map $\hat{P}(x,y,c)\in [0,1]^{H \times W \times C}$, where the most probable class for each pixel $(x,y)$ is given by the maximum value on the third dimension $c$.
An entropy map can be calculated from $\hat{P}(x,y,c)$ as follows:
\begin{eqnarray}
    \label{eq:Entropy_map}
    \hat{E}_{\mathbf{x}_i}(x,y)=\frac{-1}{\log(C)}\sum_{c=1}^C{\hat{P}(x,y,c)\log \hat{P}(x,y,c)}.
\end{eqnarray}

An entropy loss function for an image $\mathbf{x}_i$ can then be defined by adding all values of its entropy map \cite{vu_advent_2019}:
\begin{eqnarray}
    \label{eq:Entropy_loss}
    L_{ent}(\mathbf{x}_i)=\frac{1}{W H}\sum_{x,y} \hat{E}(x,y),
\end{eqnarray}
where $W$ and $H$ are width and height dimensions of $\hat{E}(x,y)$.

The network is then training with labeled synthetic images to minimize some segmentation loss and with unlabeled real images to minimize this entropy loss (see Eq.~\eqref{eq:Entropy_loss}).
In this way, the network is trained to learn the object segmentation from the synthetic images and, at the same time, it is forced to keep a low entropy on real images.
\subsection{UDA by Maximum Squares Loss}
\label{Sec:MSL}
Chen et al.\,\cite{chen_domain_2019} have observed that the gradient's magnitude of $L_{ent}$ (Eq.~\eqref{eq:Entropy_loss}) increases almost linearly with $\hat{P}(x,y,c)$ until a probability value of $\approx 0.85$ and then it grows up faster tending to infinity for a probability value of one.
As a consequence, those classes $c$ that are predicted with high probability values, dominate the training process. However, these classes -- detected with high probability -- are normally the classes that are easy to transfer.
Chen et al. give this problem the name of probability imbalance.
As a solution, they propose to change the entropy loss function by the maximum squares loss (MSL) defined as:
\begin{eqnarray}
    \label{eq:max_square_loss}
    L_{ms}(\mathbf{x}_i)=-\frac{1}{W H}\sum_c\sum_{x,y} (\hat{P}(x,y,c))^2,
\end{eqnarray}
where $W$ and $H$ are width and height dimensions of $\hat{P}(x,y,c)$.

$ L_{ms}$ has a linearly increasing gradient over the entire range of $\hat{P}(x,y,c)$ values.
By segmentation tasks, this latter prevents high confidence areas from producing excessive gradients.
Of course, these areas still have larger gradients than those with lower confidence, but their dominance is reduced in favor of the areas with lower probability and, therefore, in favor of those classes that are more difficult to transfer.

\section{\uppercase{UDA Methods for CenterNet}}

The UDA methods described in previous sections \ref{Sec:UDA} and \ref{Sec:MSL} were mainly developed for segmentation architectures.
In this work, we propose to extend these UDA methods to the anchorless detector CenterNet.
Our approach is based on the similarity between a segmentation map $\hat{P}(x,y,c)$ and the heatmap $\hat{Y}(x,y,c)$, as given by CenterNet.


\subsection{EM-Extended CenterNet}

\begin{figure*}[!ht]
    \centering
    \includegraphics[width=.95\textwidth]{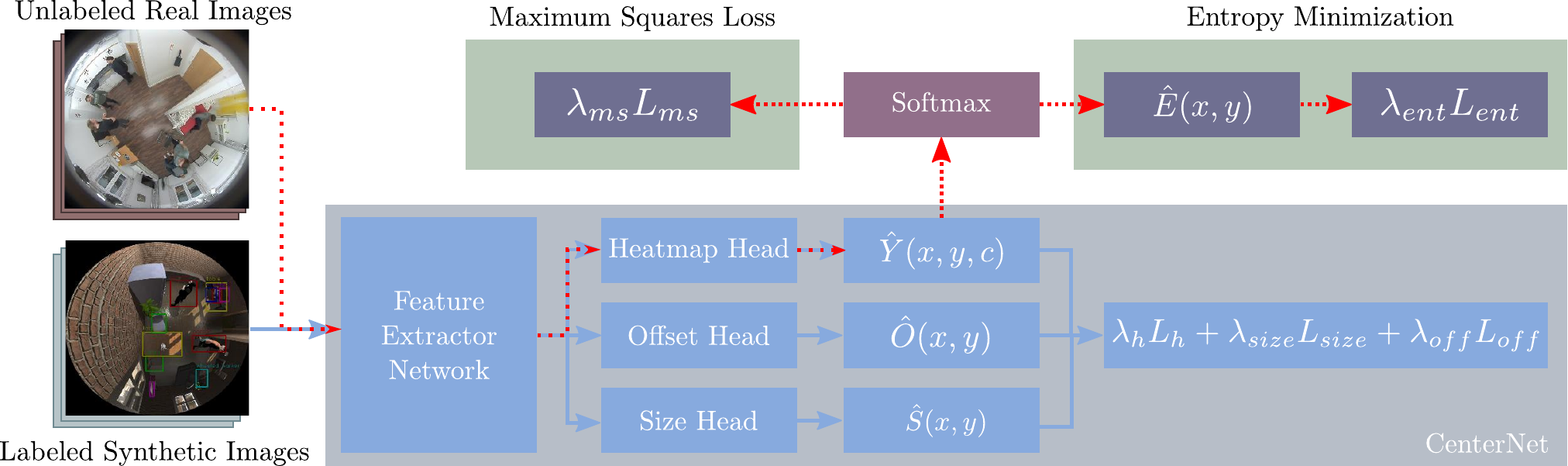}
    \caption{Extended CenterNet architecture including unsupervised domain adaptation.  }
    \label{fig:Architecture}
\end{figure*}

Equation\,\ref{eq:Entropy_map} can be used to calculate the entropy map from the CenterNet's heatmap instead from a segmentation map.
The information contained in both maps is, however, different.
A segmentation map $\hat{P}(x,y,c)$ shows the probability that a pixel $(x,y)$ belongs to a particular object class $c$.
The heatmap $\hat{Y}(x,y,c)$ shows the probability of pixel $(x,y)$ to be the center of an object of class $c$. Nevertheless, a tendency to produce low-entropy and high-entropy predictions on respectively the source and the target domain, can also be observed with CenterNet.
Figure \ref{fig:HeatmapEntropy} shows the heatmaps generated by CenterNet and their calculated entropy maps for a synthetic and a real image with class $c$ corresponding to persons.
The center points detected on the synthetic image are more defined than those on real one, and therefore present a lower entropy.

To introduce entropy minimization (EM) for domain adaptation into CenterNet, we first need to calculate the entropy map $\hat{E}(x,y)$ from the heatmap $\hat{Y}(x,y,c)$ and then to include the entropy loss, as defined in Eq.~\eqref{eq:Entropy_loss}.
To calculate the entropy from the heatmap $\hat{Y}(x,y,c)$, predicted by CenterNet, this one must first go through a Softmax function.
The resulting heatmap $Y'(x,y,c)=\text{Softmax}\{\hat{Y}(x,y,c)\}$ ensures that the sum of all center predictions along the dimension $c$ is equal to one.
The calculation of the entropy map for an image $\mathbf{x}_i$ is then performed by replacing the segmentation map $\hat{P}(x,y,c)$ with $Y'(x,y,c)$ in Eq.~\eqref{eq:Entropy_map} leading to:

\begin{eqnarray}
    \label{eq:Entropy_map_CN}
    \hat{E}_{\mathbf{x}_i}(x,y)=\frac{-1}{\log(C)}\sum_{c=1}^C{Y'(x,y,c)\log Y'(x,y,c)}.
\end{eqnarray}

Figure \ref{fig:Architecture} shows a schematic of the extended CenterNet architecture including EM.
The training set contains now a set of labeled synthetic images and a set of unlabeled real images.
The labeled synthetic images follow the blue continuous lines to contribute to the detection loss function $L_{det}$ (Eq.~\eqref{eq:detection_loss}).
The real images, on the other hand,  follow the red discontinuous lines going only through the heatmap head and contributing to the entropy loss function $L_{ent}$ as defined in Eq.~\eqref{eq:Entropy_loss} times a scale factor $\lambda_{ent}$. The final loss function for a given training image $\mathbf{x}_i$ is given by:
\begin{eqnarray}
    \label{eq:final_loss}
    L(\mathbf{x}_i)=L_{det}(\mathbf{x}_i)+\lambda_{ent}L_{ent}(\mathbf{x}_i).
\end{eqnarray}

\subsection{MSL-Exteded CenterNet}

To implement the maximum squares loss (MSL) in CenterNet, it is only necessary to replace $\hat{P}(x,y,c)$ in Eq.~\eqref{eq:max_square_loss} by $Y'(x,y,c)=\text{Softmax}\{\hat{Y}(x,y,c)\}$ obtaining:
\begin{eqnarray}
    \label{eq:max_square_loss_CN}
    L_{ms}(\mathbf{x}_i)=-\frac{R}{W H}\sum_c\sum_{x,y} (Y'(x,y,c))^2.
\end{eqnarray}
Figure \ref{fig:Architecture} also shows the extended CenterNet architecture including MSL.
Similar to EM, the labeled synthetic images contribute to the detection loss function $L_{det}$ (Eq.~\eqref{eq:detection_loss}) and the unlabeled real images to the maximum squares loss as defined in Eq.~\eqref{eq:max_square_loss_CN} time a scale factor $\lambda_{ms}$.

The final loss function for a given training image $\mathbf{x}_i$ is then given by:
\begin{eqnarray}
    \label{eq:final_loss_msl}
    L(\mathbf{x}_i)=L_{det}(\mathbf{x}_i)+\lambda_{ms}L_{ms}(\mathbf{x}_i).
\end{eqnarray}
\section{\uppercase{Results}}

\begin{table*}[!ht]
    \vspace{0.15cm}
    \caption{Results for anchor-based methods, CenterNet and CenterNet with unsupervised domain adaptation.}\label{tab:results1} \centering
    \begin{tabular}{l|cccccc|c}
        \toprule
        Class $AP_{50}$                          & Armchair       & Chair          & Person         & Table          & TV             & Walker         & mAP            \\
        \midrule
        SSD \cite{scheck_learning_2020}          & 0.021          & 0.231          & 0.904          & 0.824          & 0.545          & 0.623          & 0.525          \\
        R-FCN \cite{scheck_learning_2020}        & 0.262          & 0.039          & 0.849          & 0.859          & 0.000          & 0.640          & 0.442          \\
        Faster R-CNN \cite{scheck_learning_2020} & 0.148          & 0.141          & 0.873          & \textbf{0.980} & 0.943          & 0.613          & 0.616          \\
        \midrule
        CenterNet/DLA34                          & 0.162          & 0.451          & 0.886          & 0.949          & 0.940          & 0.656          & 0.674          \\
        CenterNet/ResNet101                      & 0.261          & 0.148          & 0.818          & 0.839          & 0.933          & 0.682          & 0.613          \\
        \midrule
        CenterNet/DLA34 EM                       & 0.202          & \textbf{0.525} & \textbf{0.912} & 0.854          & 0.967          & 0.679          & \textbf{0.690} \\
        CenterNet/ResNet101 EM                   & \textbf{0.285} & 0.382          & 0.849          & 0.915          & 0.953          & 0.648          & 0.672          \\
        CenterNet/DLA34 MSL                      & 0.106          & 0.476          & 0.857          & 0.924          & 0.937          & \textbf{0.712} & 0.668          \\
        CenterNet/ResNet101 MSL                  & 0.244          & 0.490          & 0.867          & 0.873          & \textbf{0.990} & 0.676          & \textbf{0.690} \\
        \bottomrule
    \end{tabular}
    \label{tb:mAP_Results}
\end{table*}

\subsection{Implementation}
For our experiments we use three datasets of indoor scenes captured with a top-view omnidirectional camera.
This kind of datasets are tipical for AAL (Ambient Assisted Living) applications and are a good example of the need and advantages of synthetic images, since there is no dataset with a sufficiently large number of real omnidirectional images and their corresponding labels for AAL.

In this work, we use THEODORE \cite{scheck_learning_2020} as synthetic (source) dataset, CEPDOF \cite{duan_rapid_2020} as unlabeled real (target) dataset and finally FES (Fisheye Evaluation Suite)\cite{scheck_learning_2020} as test dataset.
THEODORE and FES are annotated datasets with $C=6$ object classes: armchair, TV, table, chair, person and walker.
THEODORE has 25,000 synthetic images, whereas FES has 301 real images.
On the other hand, CEPDOF was developed for the network RAIPD \cite{duan_rapid_2020}, which is designed for omnidirectional images.
CEPDOF has 25,000 annotated frames with rotated bounding boxes only for the class person.
In our work, we used CEPDOF for the unsupervised domain adaptation such that these notations are neither required nor relevant.
We test with two feature extractors for CenterNet: ResNet101 and DLA34.
For DLA34, we include, as suggested in \cite{zhou_objects_2019}, deformable convolutions before upsampling with transposed convolutions.
In the case of ResNet101, for convenience in the implementation, no deformable convolutions are considered.

The input size $H\times W$ of CenterNet is $512\times 512$ pixels for training and $800\times800$ for testing.
In addition, we use a scale factor $R=4$ that gives a heatmap of $128 \times 128 \times 6$ pixels for training.
The focal loss function \cite{lin_focal_2017} used in CenterNet for the heatmap requires two parameter $\alpha=2$ and $\beta=4$.
The scale factors are selected as follows: $\lambda_h=1$, $\lambda_{off}=1$, $\lambda_{size}=0.1$, $\lambda_{ent}=0.0001$ and $\lambda_{ms}=0.3$.
The threshold to validate a peak as a detected center of an object is given by $\hat{Y}(x,y,c)\geq 0.1$.

For training we use Adam optimizer, learning rate 0.0001, weight decay 0.0001, manual step decay at epoch 30 with a gamma factor of 0.1 and batch sizes of 16 for each domain.
As augmentation techniques we use: flipping, rotation, translation, scaling, cropping, motion blur, adding Gaussian noise and changing  hue and brightness.

Each experiment is repeated with and without a UDA method.
The experiments without UDA allow us to evaluate the performance of the anchorless network CenterNet by direct transfer, i.e., when it is trained only with synthetic images (THEODORE) and tested on real ones (FES).
The results are compared with those presented in (Scheck et al., 2020), which were obtained based on anchor-based detectors on the same datasets that we use.
Finally, all experiments are repeated, but now incorporating the described UDA methods.
This allows evaluating whether the detection performance on real images improves by the proposed approaches in this paper.

%

\begin{figure*}[!ht]
    \centering
    \includegraphics[width=1\textwidth]{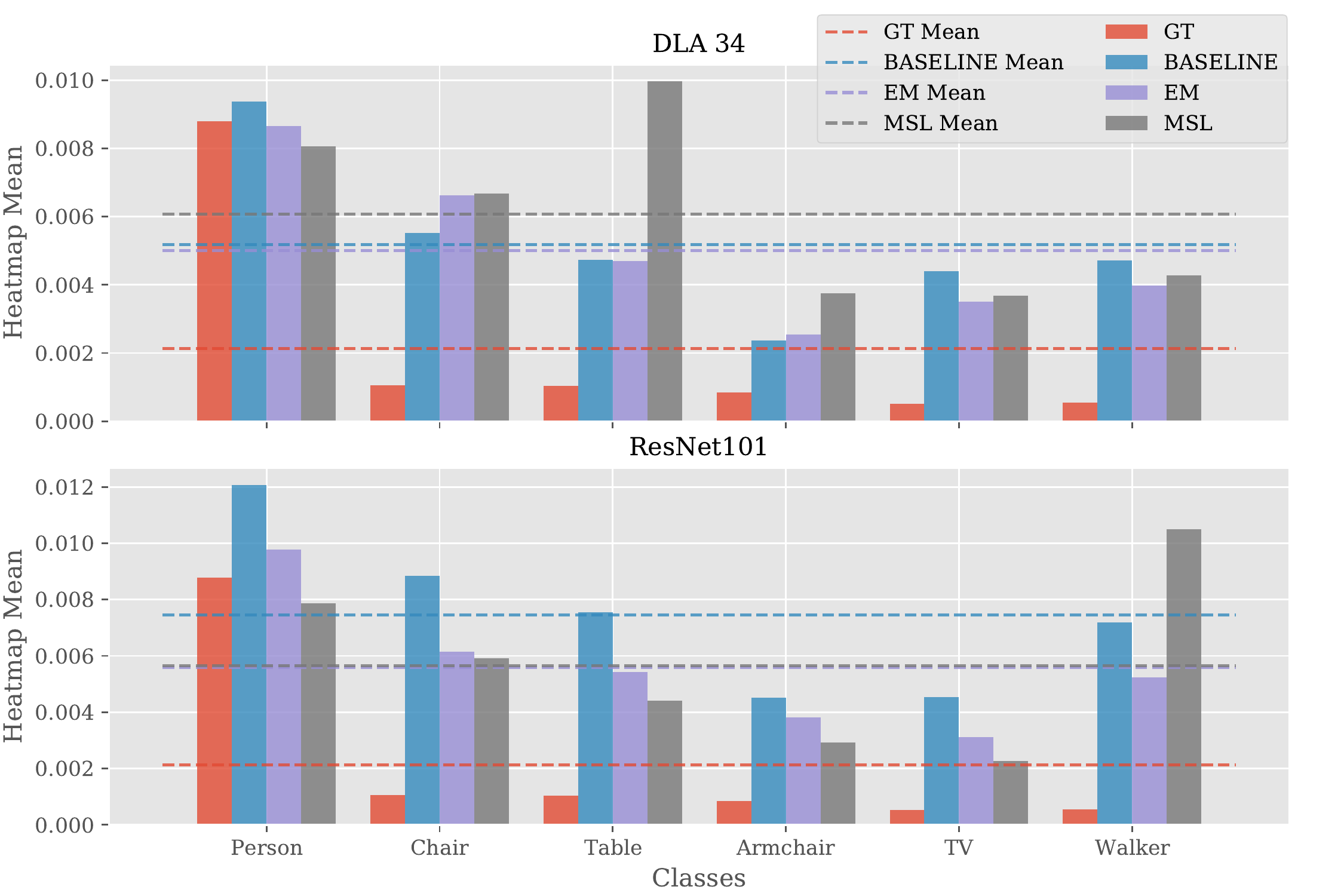}
    \caption{
        The bars show, for both feature extractors, the mean values of the heatmaps obtained from the ground truths (GT) and by applying direct transfer (baseline) and the proposed UDA approaches.
        The GTs represent the optimal heatmaps generated from the centers and size of the given bounded boxes.}
    \label{fig:results}
\end{figure*}

\subsection {CenterNet vs. Anchor-based Detectors without UDA}
As mentioned above, our first experiment consists in comparing the performance of (the anchorless) CenterNet and of anchor-based methods when applying direct transfer.

The work in \cite{scheck_learning_2020} presents the results of three anchor-based detectors, SSD \cite{fu_dssd_2017}, R-FCN \cite{dai_r-fcn:_2016} and Faster R-CNN \cite{ren_faster_2017} by training on THEODORE and testing on FES.
We repeat the same experiments using CenterNet/DLA34 and CenterNet/ResNet101 without domain adaption.

Table \ref{tb:mAP_Results} shows the resulting Average Precision (AP) per class and the final mean Average Precision (mAP) for each architecture.
CenterNet/DLA34 presents a improvement on the mAP value with respect to the anchor-based methods.
However, this improvement is specially dominated by the class chair.
This class together with the class armchair are underrepresented in THEODORE, where they are both characterized by a single 3D-mesh, i.e., there is only one armchair and one chair model in the entire dataset.
As a consequence, these two classes give in all experiments very low AP results.
Although the results of CenterNet and anchor-based methods are similar, the performance of the first one is superior given its better time performance. 
Table \ref{tb:performance} shows the frames per second (FPS) achieved by testing with each architecture on a Nvidia GeForce RTX 2080 TI and using an input resolution of $640\times 640$.
CenterNet is significantly faster than the anchor-based networks, achieving with DLA34 almost twice as many FPS as SSD.

\begin{table}[!h]
    \caption{Testing speed for the different architectures on a Nvidia GeForce RTX 2080 TI and using an input resolution of $640\times 640$.}\label{tab:results2} \centering
    \begin{tabular}{lc}
        \toprule
        Architecture                             & FPS         \\
        \midrule
        SSD \cite{scheck_learning_2020}          & 25          \\
        R-FCN \cite{scheck_learning_2020}        & 22          \\
        Faster R-CNN \cite{scheck_learning_2020} & 24          \\
        \midrule
        CenterNet/DLA34                          & \textbf{49} \\
        CenterNet/ResNet101                      & 39          \\
        \bottomrule
    \end{tabular}
    \label{tb:performance}
\end{table}

\subsection {UDA-Extended CenterNet}
As shown in Fig.~\ref{fig:HeatmapEntropy}, the heatmap $\hat{Y}(x,y,c)$ predicted by CenterNet via direct transfer presents object's centers, that are more defined or compact on synthetic images (source images) than on real ones (target images).
By incorporating unlabeled real images during the training and extendeding CenterNet to include the loss functions in Eq.~\eqref{eq:final_loss} and Eq.~\eqref{eq:final_loss_msl}, the network is forced to increase the compactness of the detected centers also on real images.
This latter can be visualized by comparing the mean value of the resulting heatmaps with and without domain adaptation. Figure \ref{fig:results} presents the heatmaps' mean values for CenterNet/DLA34 and CenterNet/ResNet101 for EM and MSL as well as for direct transfer (Baseline).
Additionally, also the mean values of the ground truth (GT) heatmaps are shown for comparison.
In the case of CenterNet/ResNet101, both proposed methods, EM ad MSL, achieve a reduction on the heatmaps' mean values.
For CenterNet/DLA34 only the EM method obtained a lower mean value as the baseline.

Table \ref{tb:mAP_Results} shows the average precision (AP) values for CenterNet/ResNet101 and CenterNet/DLA34 after applying EM and MSL for all classes.
We can observe that, except for the class table, all other AP values have increased by applying one of the proposed UDA methods.
The best mAP result is given by CenterNet/DLA34 EM and CenterNet/ResNet101 MSL, where the mAP values increase with reference to the baseline from $0.674$ to $0.69$ and from $0.613$ to $0.69$ respectively.

This means also an increment of the mAP value with respect to anchor-based methods from $0.616$ (achieved by Faster R-CNN) to $0.69$ (achieved by CenterNet/DLA34 EM and CenterNet/ResNet101 MSL).
Moreover, as mentioned before, CenterNet/DLA34 is almost twice as fast as all tested anchor-based architectures (Table \ref{tb:performance}).

\section{\uppercase{Conclusions}}
In this work, we extended two unsupervised domain adaptation (UDA) methods to an anchorless object detector, viz., CenterNet.
We consider omnidirectional synthetic images as source domain and omnidirectional real images as target domain.
Taking advantage of the CenterNet's architecture, we adapted two segmentation UDA methods, namely, minimization entropy (EM) and maximum squares loss (MSL), to the case of object detection.
Our results show that the performance of CenterNet obtained via direct transfer can be improved by applying the proposed UDA methods.
This latter validates the use of ME and MSL in order to reduce the gap between source and target domain for object detection in an anchorless architecture as illustrated for the case CenterNet.
The proposed method also enjoys the speed advantage of the anchorless detectors, being up to twice as fast as the anchor-based methods.
As future work, we plan to test our UDA methods with other anchorless detectors, adversarial approaches and including other image datasets.
\section*{\uppercase{Acknowledgements}}
\noindent This work is funded by the European Regional Development Fund (ERDF) under the grant number 100-241-945.
\pagebreak
\bibliographystyle{apalike}
{\small
    \bibliography{Paper_Vissap21}}

\end{document}